\documentclass[letterpaper,english,conference]{IEEEtran}
\usepackage[T1]{fontenc}
\usepackage[latin9]{inputenc}
\usepackage{xcolor}
\usepackage{array}
\usepackage{verbatim}
\usepackage{multirow}
\usepackage{amsmath}
\usepackage{amssymb}
\usepackage{graphicx}
\usepackage{microtype}

\makeatletter

\pdfpageheight\paperheight
\pdfpagewidth\paperwidth

\providecommand{\tabularnewline}{\\}

\usepackage{times}
\usepackage{epsfig}
\usepackage{graphicx}
\usepackage{flushend}

\usepackage{microtype}

\makeatother

\usepackage{babel}
\begin{document}
\title{Object-Centric Representation Learning for Video Question Answering}
\author{Long Hoang Dang, Thao Minh Le, Vuong Le, Truyen Tran\\
Applied Artificial Intelligence Institute, Deakin University, Australia\\
 \texttt{\small{}\{hldang,lethao,vuong.le,truyen.tran\}@deakin.edu.au}}
\maketitle
\begin{abstract}
Video question answering (Video QA) presents a powerful testbed for
human-like intelligent behaviors. The task demands new capabilities
to integrate video processing, language understanding, binding abstract
linguistic concepts to concrete visual artifacts, and deliberative
reasoning over space-time. Neural networks offer a promising approach
to reach this potential through learning from examples rather than
handcrafting features and rules. However, neural networks are predominantly
feature-based \textendash{} they map data to unstructured vectorial
representation and thus can fall into the trap of exploiting shortcuts
through surface statistics instead of true systematic reasoning seen
in symbolic systems. To tackle this issue, we advocate for object-centric
representation as a basis for constructing spatio-temporal structures
from videos, essentially bridging the semantic gap between low-level
pattern recognition and high-level symbolic algebra. To this end,
we propose a new query-guided representation framework to turn a video
into an evolving relational graph of objects, whose features and interactions
are dynamically and conditionally inferred. The object lives are then
summarized into r\a'esum\a'es, lending naturally for deliberative relational
reasoning that produces an answer to the query. The framework is evaluated
on major Video QA datasets, demonstrating clear benefits of the object-centric
approach to video reasoning.

\end{abstract}

\global\long\def\vb{\boldsymbol{v}}%
 
\global\long\def\qb{\boldsymbol{q}}%
 
\global\long\def\cb{\boldsymbol{c}}%
\global\long\def\rb{\boldsymbol{r}}%

\global\long\def\softmax{\mathrm{softmax}}%

\global\long\def\sigmoid{\mathrm{sigmoid}}%

\global\long\def\softplus{\mathrm{softplus}}%

\section{Introduction}

Answering natural questions regarding a video suggests the capability
to purposefully reason about what we see in a dynamic scene. Modern
neural networks promise a scalable approach to train such reasoning
systems directly from examples in the form (video, question, answer).
However, the networks' high degree of trainability leads to an undesirable
behavior: They tend to exploit shallow patterns and thus creating
shortcuts through surface statistics instead of true systematic reasoning\emph{.
}Stretching this methodology leads to\emph{ }data inefficiency, poor
deliberative reasoning\emph{, }and limited systematic generalization
to novel questions and scenes \cite{greff2020binding}.

\begin{figure}[th]
\includegraphics[width=1\columnwidth]{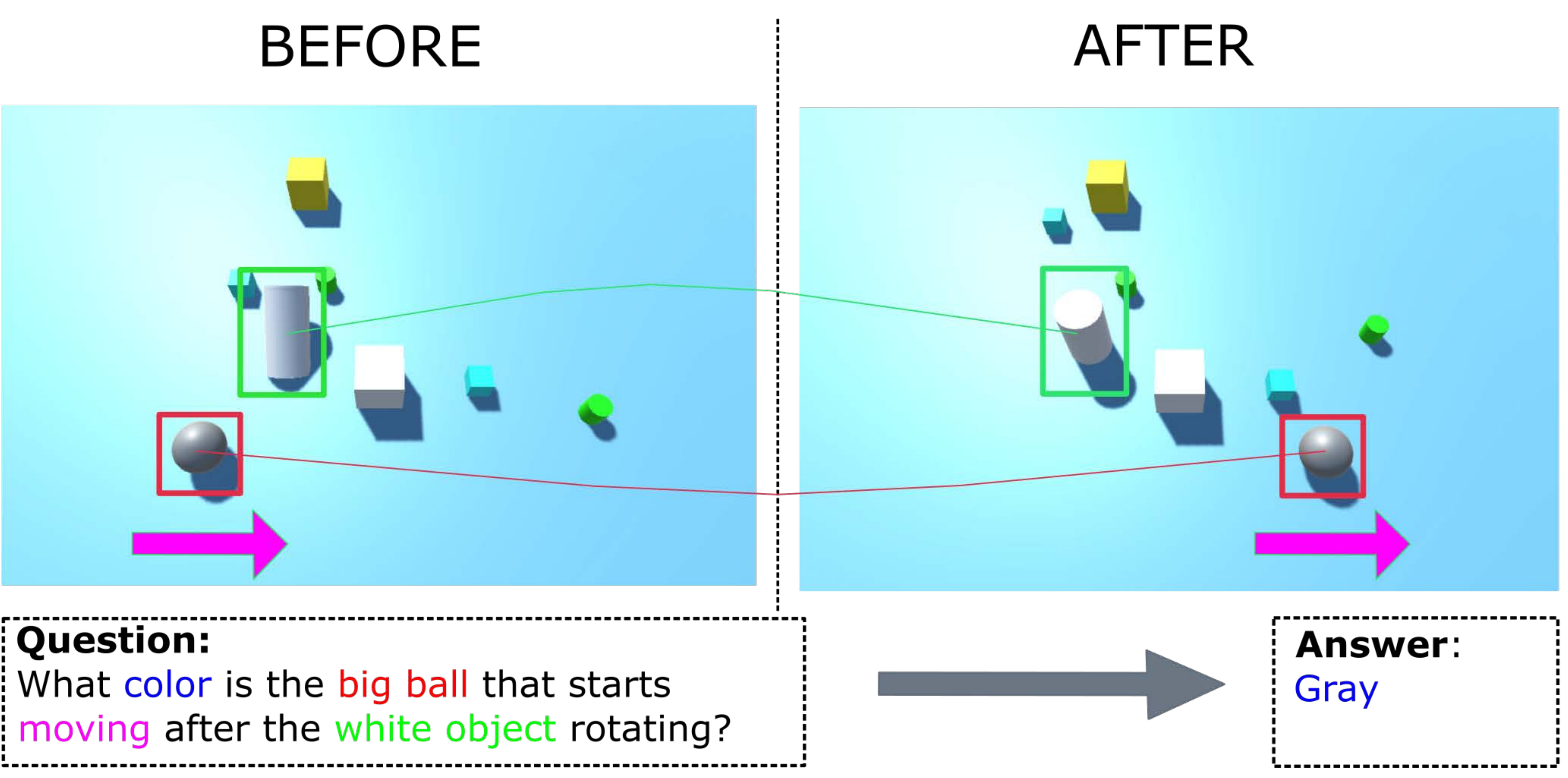}

\caption{The\emph{ }\textcolor{red}{\emph{big ball}} can have two overall temporal
parts, one before \textcolor{purple}{\emph{moving}}, and the other
after. To predict a correct answer, the model needs to localize the
action (\textcolor{purple}{\emph{moving}}), understand the evolution
of the object (\emph{the }\textcolor{red}{\emph{big ball}}) and capture
the contextualized interaction with its neighbors (\emph{after the}\textcolor{lime}{\emph{
write object}}\emph{ rotating}). \label{fig:The-big-ball}}
\end{figure}

Humans take a different approach. Since early ages, we identify objects
as the core ``living'' construct that lends itself naturally to
high-level whole scene reasoning of object compositions, temporal
dynamics, and interaction across space-time \cite{spelke2007core}.
Objects admit spatio-temporal principles of cohesion, continuity in
space-time, and local interactions, allowing humans to infer the past
and predict the future without relying too much on constant sensing.
Cognitively, objects offer a basis for important abstract concepts
such as sets and numbers, which can be symbolically manipulated without
grounding to sensory modalities; and concrete concepts such as spatial
layout and affordances. Mathematically, objects offer a decompositional
scheme to disentangle the complexity of the world, giving rise to
modularity through encapsulation, separating sensory patterns from
functions, thus enabling high-order transfer learning across completely
separate domains. 

\begin{figure*}[t]
\begin{centering}
\includegraphics[width=0.95\textwidth]{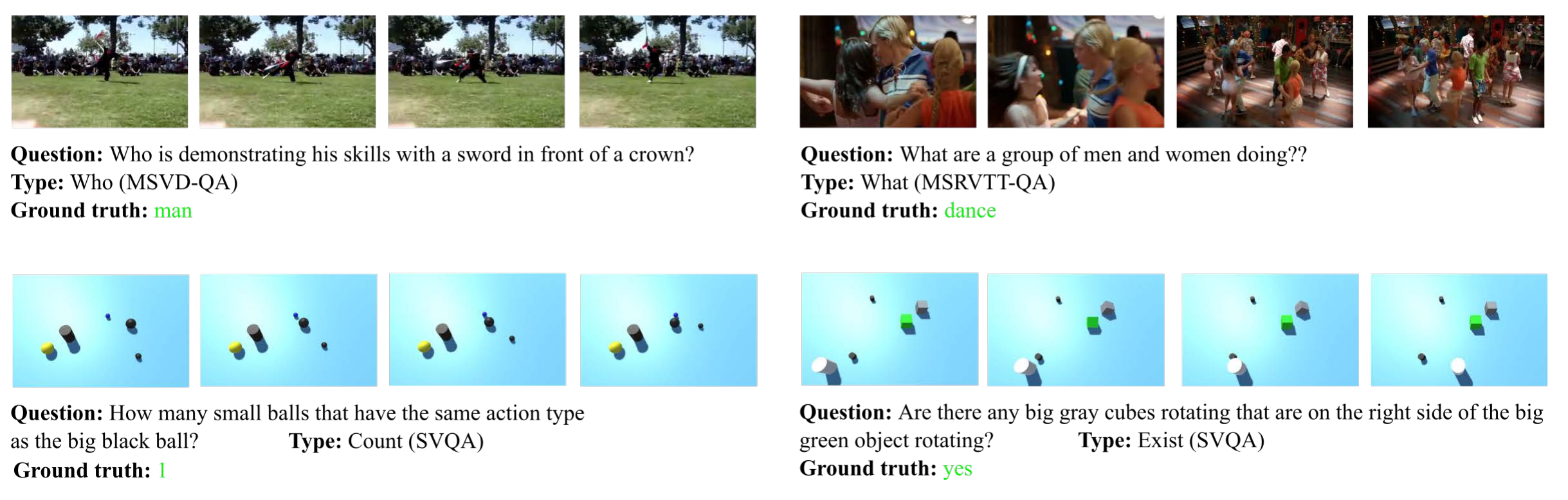}
\par\end{centering}
\caption{Several examples of scenes and question-answer pairs in Video QA datasets
with a wide range of question types. It is clear to see that Video
QA models must understand the compositions, temporal dynamics, and
interaction across space-time of objects in a whole video to provide
the correct answer. \label{fig:quanlitative_result}}
\end{figure*}

Inspired by these capabilities, we advocate for new paths to train
neural networks for video question answering (Video QA) via \emph{object-centric}
\emph{video representation} \cite{desta2018object}. Here objects
in video are primary constructs that have unique evolving lives throughout
space-time. In addition to usual visual parts, moving objects have
temporal parts \cite{hawley2020temporal}, and these are essential
to understand its evolution and contextualized interaction with other
objects. More concretely, we first extract from a video as a set of
object tubelets using recent deep neural nets for object detection
and tracking. Tubelets are then partitioned into short sub-tubelets,
each of which corresponds to a brief period of object life with small
change in appearance and position. This allows object representations
to be temporally summarized and refined, keeping only information
relevant to the query. Objects living in the same period form context-specific
relationships, which are conditionally inferred from the scene under
the guidance of the query. The objects and their relationships are
represented as a graph, one per period. As the objects change throughout
the video, their relationships also evolve accordingly. Thus the video
is abstracted as a \emph{query-conditioned evolving object graph}.

The object graphs are then parameterized as a sequence of deep graph
convolutional networks (DGCNs) \cite{Kipf2017SemiSupervisedCW}. The
DGCNs refine the object's temporal part representations within each
short period. The representation serves as input to BiLSTM, which
sequentially connects different temporal parts of the same object.
Thus the first and the last states of a BiLSTM effectively encode
the lifetime information of an object in the context of others, i.e.,
a contextualized r\a'esum\a'e. At this stage, the video is abstracted into
a set of r\a'esum\a'es which are then reasoned about given a linguistic
query using any general-purpose relational reasoning engine.

The system is evaluated on three major Video QA datasets: MSVD-QA
\cite{xu2017video} consisting of 50K questions with 2K videos, MSRVTT-QA
\cite{Xu2016MSRVTTAL} with 243K questions over 10K videos, and SVQA
\cite{emrvqasongMM18} with 120K questions over 12K videos of moving
objects. These datasets are suitable to test the reasoning capability
against complex compositional questions about spatial-temporal relationships
among objects. Our results establish new state-of-the-art accuracies
on the datasets.

To summarize, we make the following contributions: (a) proposal of
a new object-centric representation for Video QA, (b) introduction
of a new dynamic graph neural architecture that enables learning to
reason over multiple objects, their relations and events, and (c)
establishing new state-of-the-art results on three common Video QA
benchmarks of MSVD-QA, MSRVTT-QA, and SVQA.

\section{Related Work \label{sec:Related-Work}}

\begin{figure*}[t]
\begin{centering}
\includegraphics[width=0.65\paperwidth]{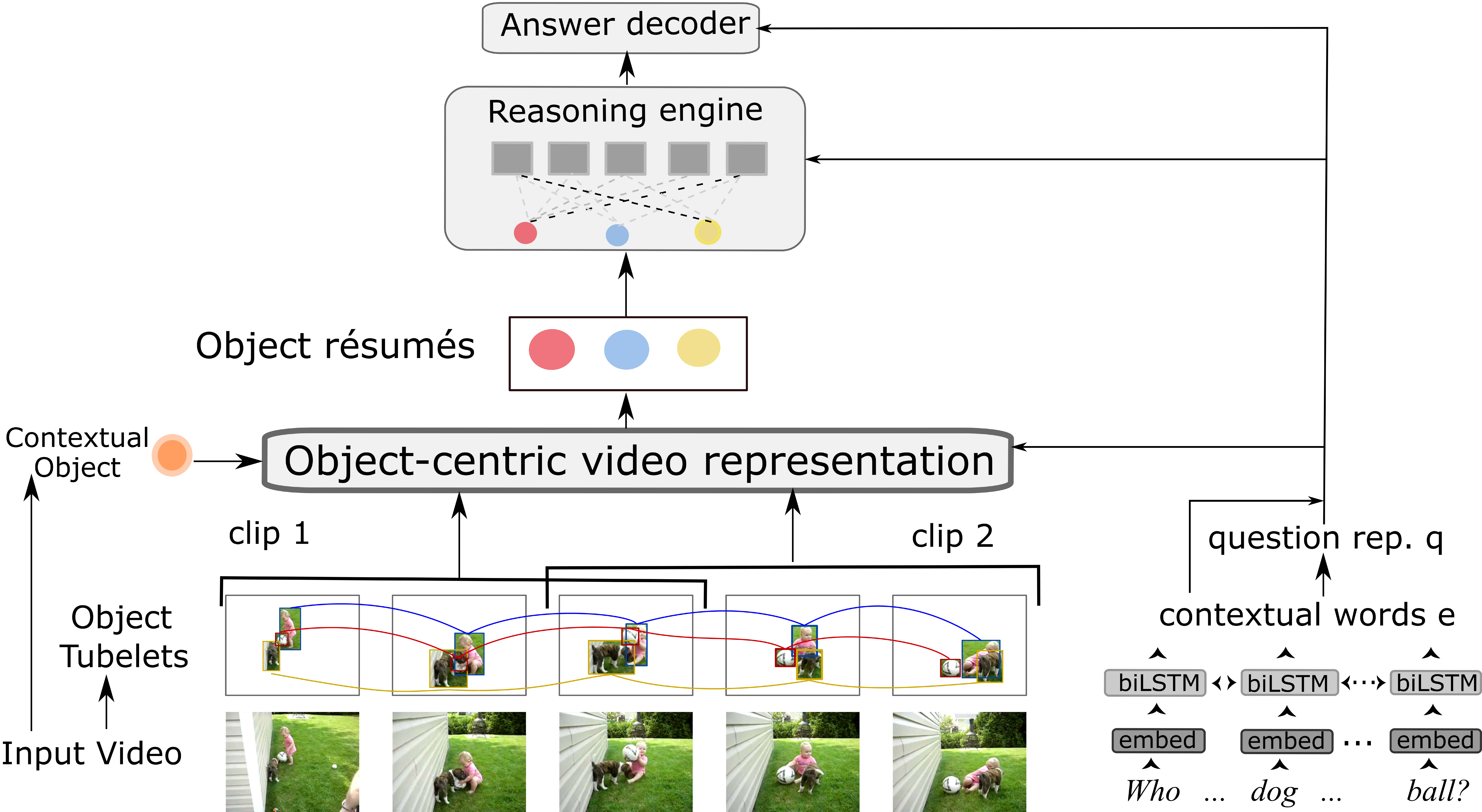}\vspace{-3mm}
\par\end{centering}
\caption{The overall architecture of our proposed model. We represent a \emph{object-centric
video representation} where a video is abstracted as a dynamic set
of interacting objects living in space-time. The object-centric video
representation takes as input tubelets of objects tracked through
time and the linguistic query to model the interactions of objects
under the modulation of the linguistic interpretation. Output of the
object-centric video representation is a set of object r\a'esum\a'es. See
Fig. \ref{fig:rel-temp-graph} for detailed design of this module.
These object r\a'esum\a'es then serves as the knowledge base for a general-purpose
reasoning engine to retrieve the relevant visual information to the
query. Finally, an answer decoder translate the presentation to an
answer prediction.\label{fig:Model-architecture}}
\end{figure*}

\emph{Question answering} in its fullest form requires a strong reasoning
capability for deliberately manipulating previously acquired knowledge
to form the answer of a given question \cite{bottou2014machine}.
While high-level reasoning in humans seems to involve symbols and
logic \cite{marcus2018algebraic}, the computational substrate is
largely neuronal, suggesting that there exists an intermediate form
of the reasoning process that is not yet formally logical, but still
powerful \cite{bottou2014machine}. Neural networks can be a good
computational model for reasoning \cite{greff2020binding,le2020dynamic},
but networks need to be dynamic and deliberative as driven by the
question, stepping away from simple associative responses. The dynamics
can be manifested in many ways: By rearranging computational units
on-the-fly \cite{hu2017learning}, by constructing relations as they
emerge and linking them to the answer \cite{le2020dynamic}, or by
dynamically building neural programs from memories \cite{le2020neural}.

\emph{Visual reasoning} presents a great challenge because the formal
knowledge needed for reasoning has not been acquired, but must be
inferred from low-level pixels. Adding to this challenge, visual question
answering typically involves high-level symbolic cues presented in
the form of linguistic questions, thus requiring automatic binding
between linguistic symbols and visual concepts \cite{hu2017learning,le2020dynamic}.
A powerful way to bridge this semantic gap is through \emph{object-centric
representation} as an abstraction over pixels \cite{greff2020binding,le2020dynamic}.
Each object has the potential to package object instance features
such as its spatio-temporal information, shape, and color, which can
solve the significant problem of disentangling complex scenes into
manageable units. Learning objects and relations have been demonstrated
to be essential for symbolic representations \cite{garnelo2019reconciling}.
A popular way to construct objects is through semantic segmentation
or object detection. Extending to video requires tracking algorithms
which exploit the temporal permanence of object to form tubelets \cite{kalogeiton2017action}.

In recent years\textbf{ }\emph{video question answering} (Video QA)
has become a major playground for spatio-temporal representation and
visual-linguistic integration. A popular approach treats video as
a sequential data structure to be queried upon. Video is typically
processed using classic techniques such as RNN \cite{zhu2017uncovering}
or 3D CNN \cite{qiu2017learning}, augmented by attention mechanisms
\cite{ye2017video}, or manipulated in a memory network \cite{kim2017deepstory}.
Stepping away from this flat structure, video can be abstracted as
a hierarchy of temporal relations \cite{le2020hierarchical,le2020neural-reason},
which can either directly generate answers given the query, or serve
as a representation scheme to be reasoned about by a generic reasoning
engine \cite{le2020neural-reason}. More recently, objects have been
suggested to be the core component of Video QA \cite{yi2019clevrer}
as they offer clean structural semantics compared to whole-frame unstructured
features. Our work pushes along this line of object-centric representation
for Video QA, built on the premise that object interactions are local,
and their representation should depend on other objects in context
as well as being guided by the query. In particular, objects are tracked
into tubelets so that their lifelines are well defined. Their attributes
and relations along such lifeline are constantly updated in the dynamic
graph of their social circles of co-occurred fellow objects.

\section{Method \label{sec:Method}}

The Video QA task seeks to build a conditional model $P_{\theta}$
to infer an answer $a$ from of set $A$, given a video $V$ and a
natural linguistic query $q$:
\begin{equation}
\bar{a}=\arg\max_{a\in A}P_{\theta}\left(a\mid q,V\right).\label{eq:VQA-overall}
\end{equation}

The challenges stem from (a) the long-range dependencies intra\textendash objects
across time, and inter\textendash objects across space in the video
$V$; (b) the arbitrary expression of the linguistic question $q$;
and (c) the spatio-temporal reasoning over the visual domain as guided
by the linguistic semantics. Fig. \ref{fig:quanlitative_result} gives
some typical examples that can describe the major problems in Video
QA task. In what follows, we present our object-oriented solution
to tackle these challenges.

\subsection{Model Overview \label{subsec:Overview}}

Fig\@.~\ref{fig:Model-architecture} illustrates the overall architecture
of our model. Our main contribution is the object-centric video representation
for Video QA in which a video is abstracted as a dynamic set of interacting
objects that live in space-time. The interactions between objects
are interpreted in the context information given by the query (See
Sec. \ref{subsec:Object-centric-Video-Representat}). The output of
the object-centric video representation is a set of object r\a'esum\a'es.
These r\a'esum\a'es later serve as a knowledge base for a general-purpose
relational reasoning engine to extract the relevant visual information
to the question. In the reasoning process, it is desirable that query-specific
object relations are discovered and query words are bound to objects
in a deliberative manner. The reasoning module can be generic, as
long as it can handle the natural querying over the set of r\a'esum\a'es.
At the end of our model, an answer decoder takes as input the output
of the reasoning engine and the question representation to output
a vector of probabilities across words in the vocabulary for answer
prediction.

\paragraph*{Linguistic question representation}

We make use of BiLSTM with GloVe word embedding \cite{pennington2014glove}
to represent a given question $q$. In particular, each word in the
question is first embedded into a vector of 300 dimensions. A BiLSTM
running on top of this vector sequence produces a sequence of state
pairs, one for the forward pass and the other for the backward pass.
Each state pair becomes a contextual word representation $e_{s}=\left[\overleftarrow{h_{s}},\overrightarrow{h_{s}}\right],\ e_{s}\in\mathbb{R}^{d}$,
where $d$ is the vector length, and $s=1,...,S$ where $S$ is the
length of the question. The global sequential representation $q_{g}$
is obtained by combining the two end states of BiLSTM: $\qb_{g}=\left[\overleftarrow{h_{1}},\overrightarrow{h_{S}}\right],\ q_{g}\in\mathbb{R}^{d}$.
Finally, we integrate $q_{g}$ with the contextual words $e_{s}$
by an attention mechanism to output the final question representation:
\begin{equation}
q=\sum_{s=1}^{S}e_{s}*\text{softmax}_{s}(w_{q}^{\top}(e_{s}\odot q_{g})),
\end{equation}
 where $w_{q}\in\mathbb{R}^{d}$ is network learnable weights.

\subsection{Object-centric Video Representation\label{subsec:Object-centric-Video-Representat}}

\begin{figure}[t]
\begin{centering}
\includegraphics[width=1\columnwidth]{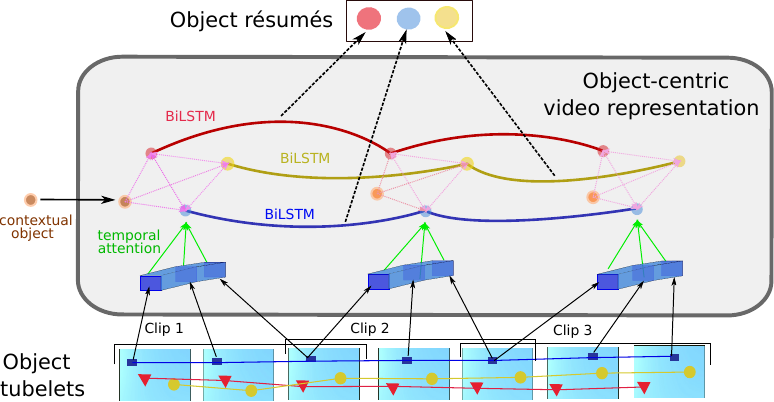}
\par\end{centering}
\caption{Object-centric Video Representation - Illustration with a scene of
three objects. The objects are first detected and tracked (thin matching
color lines) from video frames into tubelets (only drawn for the \textcolor{blue}{blue
rectangle}). Their visual and positional features are gathered on
each clip with temporal attention (\textcolor{green}{green arrows})
into temporal parts (nodes in lower row). The parts of each clip interact
by a GCN (\textcolor{purple}{dotted purple graph}) with the contextual
object (\textcolor{orange}{orange node}) and join temporally by BiLSTMs.\label{fig:rel-temp-graph}}
\end{figure}

We represent an object live based on tubelet \textendash{} the 3D
structure of the object in space-time. At each time step, an object
is a 2D bounding box with a unique identity assigned by object tracking
(Sec.~\ref{subsec:Constructing-object-tubelets}). We obtain tubelets
by simply linking bounding boxes of the same identities throughout
the given video. If seeing a video as a composition of events, the
interactions between objects often happen in a short period of time.
Hence, we can break the object tubelets into sub-tubelets where each
sub-tubelet is equivalent to a temporal part \cite{hawley2020temporal}.
As we model the interactions between objects in the context given
by the query, we then summarize each object-wise temporal part into
a query-conditioned vector representation (Sec.~\ref{subsec:Language-conditioned-representat}).
The representations of temporal parts of objects are then treated
as nodes of a \emph{query-conditioned evolving object graph} where
the edges of the graph denote the relationships of objects living
in the same temporal part.  In order to obtain the full representations
of object lives through the video, we link object representations
in consideration of their interference with neighbor objects across
temporal parts. Finally, each object life is summarized into a r\a'esum\a'e,
preparing the object system for relational reasoning. See Fig.~\ref{fig:rel-temp-graph}
for illustration of our object-centric video representation in a scene
of three objects.

\subsubsection{Constructing Object Tubelets \label{subsec:Constructing-object-tubelets}}

\paragraph*{Detecting and tracking objects}

We use Faster R-CNN \cite{ren2015faster} to detect frame-wise objects
in a given video. This returns, for each object, (a) appearance features
$\vb_{a}\in\mathbb{R}^{2048}$ (representing ``what''); (b) the
bounding box $\vb_{b}=\left[x_{min},y_{min},x_{max},y_{max}\right]$
(``where'') and (c) a confidence score. We use a popular object
tracking framework DeepSort \cite{wojke2017simple} which makes use
of the appearance features and confidence scores of detected bounding
boxes to track objects, assigning a unique ID to bounding boxes of
the same object.  For the ease of implementation, we assume that
the objects live from the beginning to the end of video and missing
objects at a time are marked with null values. Each object is now
a tubelet, consisting of a unique ID, a sequence of coordinates of
bounding boxes through time and a set of corresponding appearance
features.

\paragraph*{Joint encoding of ``what'' and ``where''}

Positions are of critical importance in reasoning about the spatial
arrangement of objects at any given time \cite{zhuang2017towards,wang2019neighbourhood}.
Therefore, to represent a geometrical information of each object tubelet
at a time step, we incorporate a spatial vector of 7-dimensions of
the bounding boxes, including their four coordinates, width and height,
and the size of their relative areas w.r.t the area of the whole video
frame, i.e $v_{p}=\left[\frac{x_{min}}{W},\frac{y_{min}}{H},\frac{x_{max}}{W},\frac{y_{max}}{H},\frac{\Delta x}{W},\frac{\Delta y}{H},\frac{\Delta x\Delta y}{WH}\right]$,
where $\Delta x=x_{max}-x_{min}$, $\Delta y=y_{max}-y_{min}$, and
$(W,H)$ are the width and height of the video frame, respectively.

Considering a brief period of time where an action takes place, the
appearance feature $v_{a}$ of an object may change very little while
the change in its position is more noticeable. For example, given
a video where a person is walking towards a car, it is likely that
the changes in position is more informative than that in appearance.
Hence, we propose to use a multiplicative gating mechanism to encode
the \emph{position-specific appearance} of objects
\begin{align}
\vb_{ap} & =f_{1}(\vb_{a})\odot g_{1}(\vb_{p}),\label{eq:pos-app-feat}
\end{align}
where $f_{1}(\vb_{a})$ is non-linear mapping function and $g_{1}(\vb_{p})\in(\boldsymbol{0},\boldsymbol{1})$
is a position gating function to take control the flow of appearance,
making the change in representation of objects at different time steps
be more discriminated from each other. We choose $f_{1}(\vb_{a})=\tanh\left(W_{a}\vb_{a}+b_{a}\right)$
and $g_{1}(\vb_{p})=\text{sigmoid}\left(W_{p}\vb_{p}+b_{p}\right)$
in our later implementation, where $W_{a}$ and $W_{p}$ are network
parameters.

Besides, in order to compensate for the loss caused by the imperfect
object detection, we incorporate a so-called\emph{ contextual object
}$v_{c}\in\mathbb{R}^{2048}$ of a video frame. Particularly, we utilize
pre-trained ResNet \cite{He2016DeepRL} and take pool5 output features
as the object feature $v_{c}$ of each time step.

\subsubsection{Language-conditioned Representation of Temporal Parts \label{subsec:Language-conditioned-representat}}

Once the \emph{position-specific appearance} of objects is computed,
we reduce the complexity by grouping several frames into a short clip,
creating a language-conditioned temporal part by taking the question
into account. The goal is to prepare the parts to be \emph{question\textendash specific},
making them more ready for subsequent relational reasoning.

We split an object life into $K$ equal temporal parts $C=\left\{ C_{1},C_{2},...,C_{K}\right\} ,$
where part $C_{k}$ contains $t$ frames. Let $C_{k}=\left\{ \vb_{ap}^{k,1},..,\vb_{ap}^{k,i},...,\vb_{ap}^{k,t}\right\} $,
where $\vb_{ap}^{k,i}$ is the position-conditioned feature computed
in Eq.~(\ref{eq:pos-app-feat}) of the $i$-th frame in the $k$-th
part. Given the question representation $\qb$, we then apply a \emph{temporal
attention} mechanism to compute the probability to which a frame
in a part is attended:
\begin{align}
\alpha_{k,i} & =\text{softmax}_{i}\left(w_{s}^{\top}\left((W_{q}\qb+b_{q})\odot(W_{v}\vb_{ap}^{k,i}+b_{v})\right)\right),\label{eq:temporal_attention}
\end{align}
where $\odot$ is the Hadamard product, $\left\{ w_{s},W_{q},W_{v}\right\} $
are learnable weights. The temporal part $k$ of an object is then
summarized as: 
\begin{equation}
\cb_{k}=\lambda\ast\sum_{i=1}^{t}\alpha_{k,i}\vb_{ap}^{k,i},\label{eq:clip-rep}
\end{equation}
where $\lambda$ is a binary mask vector to exclude missed detections
of objects. $\lambda$ is calculated based on indices of the time
steps marked with null values as in Sec.~\ref{subsec:Constructing-object-tubelets}.

\subsubsection{Query-conditioned Object Graph \label{subsec:Relating-temporal-parts}}

Imagining that we have $N$ sequences of objects living together in
a period of time (temporal part), their living paths are not only
constrained by their own behavior at different points in time but
also influenced by their object neighbors living nearby. Besides,
in the context of Video QA, relationships between objects are understood
in the context given by the query. Note that temporal object parts
living in the same period $k$ can be highly correlated, depending
on the spatial distance between objects and the semantics of the query.
We represent objects in a temporal part $k$ and their relationships
by a graph $G_{k}=(C_{k},A_{k})$, where $C_{k}=\{c_{k,o}\}_{o=1}^{N}$
are nodes and $A_{k}\in\mathbb{R}^{N\times N}$ is the adjacent matrix.
The adjacent matrix $A_{k}$ is query-dependent and is given by:
\begin{align}
a_{k,o} & =\text{softmax}_{o}(w_{a}^{\top}([\cb_{k,o},\cb_{k,o}\odot q])),\\
A_{k} & =a_{k}^{\top}a_{k},
\end{align}
where $\cb_{k,o}$ is representation of part $k$ of object $o$,
as computed in Eq.~(\ref{eq:clip-rep}), and $a_{o}$ is the probability
to which object $o$ is attended.

Given the adjacent matrix $A_{k}$, we use graph convolutional networks
(GCNs) \cite{Kipf2017SemiSupervisedCW} to correlate objects and systematically
refine all the object representations by considering their connections
with other objects in the same temporal part. We stack GCNs into layers
with skip-connections between them, allowing the refinement to run
through multiple rounds. To mitigate to effects of imperfect object
detection, we propose to use the contextual features of video frames
$v_{c}$ as explained in Sec. \ref{subsec:Constructing-object-tubelets}
as an additional node for graph $G_{k}$. Different from the original
bidirectional graph in \cite{Kipf2017SemiSupervisedCW}, our temporal-part-specific
object graph $G_{k}$ is now considered as a heterogeneous graph where
connections between objects are bidirectional while the connections
between the contextual node and other nodes are directional. Let $H_{k}^{0}=\left(\cb_{k,1},\cb_{k,2},...,\cb_{k,N}\right)\in\mathbb{R}^{N\times d}$
be the initial node embedding matrix, we refine representations of
nodes in each graph $G_{k}$ as follows:

\begin{equation}
F_{l}\left(H_{k}^{l-1}\right)=W_{2}^{l-1}g\left(A_{k}H_{k}^{l-1}W_{1}^{l-1}+W_{v_{c}}v_{c}^{k}+b^{l-1}\right),\label{eq:GCN}
\end{equation}
where $l=1,2,...L$ is the number of GCN layers, and $g\left(\cdot\right)$
is nonlinear transformation chosen to be ELU in our implementation.
The skip-connection is then added between GCN layers to do multiple
round representation refinement:
\begin{equation}
H_{k}^{l}=g\left(H_{k}^{l-1}+F_{l}\left(H_{k}^{l-1}\right)\right).\label{eq:skip_connection}
\end{equation}
Finally, the contextualized parts representation is then $\left(\tilde{\cb}_{k,1},\tilde{\cb}_{k,2},...,\tilde{\cb}_{k,N}\right)\leftarrow H_{k}^{L}$.

\subsubsection{Video as Evolving Object Graphs \label{subsec:Dynamic-object-graphs}}

An object is a sequence of temporal parts whose representation is
contextualized in Sec.~\ref{subsec:Relating-temporal-parts}. Hence
the video is abstracted as a spatio-temporal graph whose spatial and
temporal dependencies are conditioned on the query, as illustrated
in Fig.~\ref{fig:rel-temp-graph}. Let $\tilde{\cb}_{k}$ be contextualized
part representation of the object. Temporal parts are then connected
through a BiLSTM:

\begin{align}
\overrightarrow{h}_{k} & =\overrightarrow{\text{LSTM}}\left(\overrightarrow{h}_{k-1},\tilde{\cb}_{k}\right),\,k>0\,\,\,\text{and}\label{eq:clip-forward}\\
\overleftarrow{h}_{k} & =\overleftarrow{\text{LSTM}}\left(\overleftarrow{h}_{k+1},\tilde{\cb}_{k}\right),\,k<K,\label{eq:clip-backward}
\end{align}
with the first state initialized as $\overrightarrow{h}_{0}=\overrightarrow{W}_{q}\qb+\overrightarrow{b}_{q}$
and $\overleftarrow{h}_{K}=\overleftarrow{W}_{q}\qb+\overleftarrow{b}_{q}$,
respectively.

The dynamic object graph represents the scene evolution well but it
remains open how to efficiently perform arbitrary reasoning, given
the free-form expression of the linguistic query. For example, given
a scene of multiple objects of different shapes, we want to answer
a question ``Is there a cylinder that start rotating before the green
cylinder?''. For this we need to search for the matching cylinders,
and traverse the graph to identify the event of rotating, then follow
up the cylinder trajectories before the event. In the process of matching,
we need to find the right visual colored cylinder that agree with
the linguistic words ``cylinder'', ``green cylinder'' and ``rotating''.
For large temporal graphs, it poses a great challenge to learn to
perform these discrete graph operations.

To mitigate the potential complexity, we propose to integrate out
the temporal dimension of the dynamic object graph, creating an unordered
set of $N$ object r\a'esum\a'es. This method of temporal integration has
been empirically shown to be useful in Video QA \cite{le2020neural-reason}.
In particular, we compute a r\a'esum\a'e for each object by summarizing
its lifetime using $\rb=\left[\overleftarrow{h}_{1},\overrightarrow{h}_{K}\right]$,
where $\overleftarrow{h}_{1}$ and $\overrightarrow{h}_{K}$ are end
states of the BiLSTM using Eqs.~(\ref{eq:clip-forward},\ref{eq:clip-backward}).
This representation implicitly codes the appearance, geometric information,
temporal relations of an object in space-time, as well as the parts
relation between objects living in the same spatio-temporal context.

\subsection{Relational Reasoning \label{subsec:Relational-reasoning}}

Now we have two sets, one consisting of contextualized visual r\a'esum\a'es
$R=\left\{ \rb_{o}\right\} _{o=1}^{N}$, and the other consisting
of the query representation $Q=\left\{ q_{g},e_{1},e_{2},...,e_{S}\right\} $.
Reasoning, as defined in Eq.~(\ref{eq:VQA-overall}), amounts to
the process of constructing the interactions between these two sets
(e.g., see \cite{hudson2018compositional}). More precisely, it is
the process of manipulating the representation of the visual object
set, as guided by the word set. An important manipulation is to dynamically
construct \emph{language-guided predicates of object relations}, and
chain these predicates in a correct order so that the answer emerges.
It is worth to emphasize that our\emph{ object-centric} \emph{video
representation} can combine with a wide range of reasoning models.
Examples of the generic reasoning engines are the recently introduced
MACNet (Memory, Attention, and Composition Network) \cite{hudson2018compositional}
and LOGNet (Language-binding Object Graph Network) \cite{le2020dynamic}.
Although these models were originally devised for static images, its
generality allows adaptation to Video QA through object r\a'esum\a'es.

\paragraph*{Dual-process view }

The application of generic iterative relational reasoning engines,
for object-oriented Video QA somewhat resembles the dual-process idea
proposed in \cite{le2020neural-reason}. It disentangles the visual
QA system into two sub-processes \textendash{} one for domain-specific
modeling, and the other for deliberative general-purpose reasoning.

\subsection{Answer Decoder}

We follow prior works in Video QA \cite{le2020neural-reason,fan2019heterogeneous,le2020hierarchical}
to design an answer decoder of two fully connected layers followed
by the softmax function to obtain probabilities of labels for prediction.
As we treat Video QA as a multi-class classification, we use cross-entropy
loss to train the model.

\section{Experiments \label{sec:Experiments}}

\subsection{Datasets}

We evaluate our proposed architecture on the three recent benchmarks,
namely, MSVD-QA, MSRVTT-QA \cite{xu2017video,Xu2016MSRVTTAL} and
SVQA \cite{emrvqasongMM18}.

\textbf{MSVD-QA} contains 1,970 real-world video clips and 50,505
QA pairs of five question types: what, who, how, when, and where,
of which 61\% of the QA pairs is used for training, 13\% for validation
and 26\% for testing.

\textbf{MSRVTT-QA} consists of 243K QA pairs annotated from over 10K
real videos. Similar to the MSVD-QA dataset, the questions are classified
into five types: what, who, how, when, and where. The proportions
of videos in the training, testing and validation splits are 65/30/5\%,
respectively.

\textbf{SVQA} is designed specifically for multi-step reasoning in
Video QA, similar to the well-known visual reasoning benchmark with
static images CLEVR \cite{johnson2017clevr}. It contains 12K short
synthetic videos and 120K machine-generated compositional and logical
question-answer pairs covering five categories: \emph{attribute comparison,
count, query, integer comparison} and \emph{exist}. Each question
is associated with a question program. SVQA helps mitigate several
limitations of the current Video QA datasets such as language bias
and the lack of compositional logic structure. Therefore, it serves
as an excellent testbed for multi-step reasoning in space-time. We
follow prior work \cite{le2020neural-reason} to use 70\% of videos
as the training set, 20\% of videos as the test set, and the last
10\% for cross-validation.\textbf{ }

\subsection{Implementation Details}

Unless otherwise stated, each video is segmented into 10 clips of
16 consecutive frames. Faster R-CNN\footnote{https://github.com/airsplay/py-bottom-up-attention}
\cite{ren2015faster} is used to detect frame-wise objects. We take
40 tubelets per video for MSVD-QA and MSRVTT-QA while that of SVQA
is 30 simply based on empirical experience. Regarding the language
processing, we embed each word in a given question into a vector of
300 dimensions, which are initialized by pre-trained GloVe \cite{pennington2014glove}.
The default configuration of our model is with $L=6$ GCN layers for
the query-conditioned object graph (See Sec. \ref{subsec:Relating-temporal-parts}),
feature dimensions $d=512$ in all sub-networks. Regarding the general
relational reasoning engine, we use 12 reasoning steps on both MACNet
and LOGNet with default other parameters as in \cite{le2020dynamic,hudson2018compositional}.
We refer our models as OCRL+LOGNet or OCRL+MACNet, respectively, where
OCRL stands for\emph{ Object-Centric Representation Learning}. Without
mentioning explicitly, we report results of the OCRL+LOGNet as it
generally produces favorable performance over the OCRL+MACNet.

We train our models with Adam optimizer with an initial learning
rate of $10^{-4}$ and a batch size of 64. To be compatible with related
works \cite{le2020neural-reason,emrvqasongMM18,le2020hierarchical},
we use accuracy as evaluation metric for all tasks.

\subsection{Comparison against SOTAs}

We compare our proposed method against recent state-of-the-art models
on each dataset.\textbf{ MSVD-QA and MSRVTT-QA datasets:} Results
on the MSVD-QA and MSRVTT-QA datasets are presented in Table~\ref{tab:Experiments-on-MSVD-QA}.
As can be seen, our OCRL+LOGNet model consistently surpasses the
all SOTA methods. In particular, our model significantly outperforms
the most advanced model HCRN \cite{le2020hierarchical} by 2.1 absolute
points on the MSVD-QA, whereas it slightly advances the accuracy on
the MSRVTT-QA from 35.6 to 36.0. Note that the MSRVTT-QA is greatly
bigger than the MSVD-QA and it contains long videos with complex relations
between objects, so it might need more tubelets per video than the
current implementation of 40 tubelets. As for the contribution of
each model's component, we provide further ablation studies on MSVD-QA
dataset in Sec. \ref{subsec:Ablation-study}.

\textbf{}
\begin{table}
\textbf{\caption{Experiments on MSVD-QA and MSRVTT-QA datasets. \label{tab:Experiments-on-MSVD-QA}}
}
\centering{}%
\begin{tabular}{l|cc}
\hline 
\multirow{2}{*}{Model} & \multicolumn{2}{c}{Test accuracy (\%)}\tabularnewline
\cline{2-3} \cline{3-3} 
 & MSVD-QA & MSRVTT-QA\tabularnewline
\hline 
ST-VQA \cite{jang2017tgifqa} & 31.3 & 30.9\tabularnewline
Co-Mem \cite{gao2018motion} & 31.7 & 32.0\tabularnewline
AMU\cite{xu2017video} & 32.0 & 32.5\tabularnewline
HME \cite{fan2019heterogeneous} & 33.7 & 33.0\tabularnewline
HRA \cite{8451103} & 34.4 & 35.1\tabularnewline
HCRN \cite{le2020hierarchical} & 36.1 & 35.6\tabularnewline
\hline 
\textbf{OCRL+LOGNet} & \textbf{38.2} & \textbf{36.0}\tabularnewline
\hline 
\end{tabular}
\end{table}

\textbf{SVQA dataset:} Table~\ref{tab:SOTA-SVQA} shows the comparisons
between our model against the SOTA methods. Indeed, our proposed network
significantly outperforms all recent SOTA models by a large margin
(3.7 points). Specifically, we achieve the best performance in a majority
of sub-tasks (11/15), which account for 91\% of all question-answer
pairs, and slightly underperform in the other four categories. We
speculate that this is due the advantages of CRN+MACNet \cite{le2020neural-reason}
in explicitly encoding high-order temporal relations. Nevertheless,
the results clearly demonstrate the necessity of an object-centric
approach in handling a challenging problem as Video QA.

\begin{table*}
\caption{Experiments on SVQA dataset. SA(S), TA is short for TA-GRU, SA+TA
is short for SA+TA-GRU, both are from \cite{emrvqasongMM18}, STRN
is from \cite{Singh2019SpatiotemporalRR}, CRN+MAC is from \cite{le2020neural-reason}
in which MACNet is used for the reasoning process. OCRL+MA and OCRL+LO
refer to OCRL+MACNet and OCRL+LOGNet, respectively. \label{tab:SOTA-SVQA}}

\centering{}%
\begin{tabular}{l|c|c|c|c|c|c|c|c|c|c|c|c|c|c|c|c}
\hline 
\multirow{2}{*}{{\footnotesize{}Methods}} & \multirow{2}{*}{{\footnotesize{}Exist}} & \multirow{2}{*}{{\footnotesize{}Count}} & \multicolumn{3}{c|}{} & \multicolumn{5}{c|}{{\footnotesize{}Attribute Comparison}} & \multicolumn{5}{c|}{{\footnotesize{}Query}} & \multirow{2}{*}{\textbf{\footnotesize{}All}}\tabularnewline
\cline{4-16} \cline{5-16} \cline{6-16} \cline{7-16} \cline{8-16} \cline{9-16} \cline{10-16} \cline{11-16} \cline{12-16} \cline{13-16} \cline{14-16} \cline{15-16} \cline{16-16} 
 &  &  & {\footnotesize{}More} & {\footnotesize{}Equal} & {\footnotesize{}Less} & {\footnotesize{}Color} & {\footnotesize{}Size} & {\footnotesize{}Type} & {\footnotesize{}Dir} & {\footnotesize{}Shape} & {\footnotesize{}Color} & {\footnotesize{}Size} & {\footnotesize{}Type} & {\footnotesize{}Dir} & {\footnotesize{}Shape} & \tabularnewline
\hline 
{\footnotesize{}SA(S)} & {\footnotesize{}51.7} & {\footnotesize{}36.3} & {\footnotesize{}72.7} & {\footnotesize{}54.8} & {\footnotesize{}58.6} & {\footnotesize{}52.2} & {\footnotesize{}53.6} & {\footnotesize{}52.7} & {\footnotesize{}53.0} & {\footnotesize{}52.3} & {\footnotesize{}29.0} & {\footnotesize{}54.0} & {\footnotesize{}55.7} & {\footnotesize{}38.1} & {\footnotesize{}46.3} & {\footnotesize{}43.1}\tabularnewline
{\footnotesize{}TA} & {\footnotesize{}54.6} & {\footnotesize{}36.6} & {\footnotesize{}73.0} & {\footnotesize{}57.3} & {\footnotesize{}57.7} & {\footnotesize{}53.8} & {\footnotesize{}53.4} & {\footnotesize{}54.8} & {\footnotesize{}55.1} & {\footnotesize{}52.4} & {\footnotesize{}22.0} & {\footnotesize{}54.8} & {\footnotesize{}55.5} & {\footnotesize{}41.7} & {\footnotesize{}42.9} & {\footnotesize{}44.2}\tabularnewline
{\footnotesize{}SA+TA} & {\footnotesize{}52.0} & {\footnotesize{}38.2} & {\footnotesize{}74.3} & {\footnotesize{}57.7} & {\footnotesize{}61.6} & {\footnotesize{}56.0} & {\footnotesize{}55.9} & {\footnotesize{}53.4} & {\footnotesize{}57.5} & {\footnotesize{}53.0} & {\footnotesize{}23.4} & {\footnotesize{}63.3} & {\footnotesize{}62.9} & {\footnotesize{}43.2} & {\footnotesize{}41.7} & {\footnotesize{}44.9}\tabularnewline
{\footnotesize{}STRN} & {\footnotesize{}54.0} & {\footnotesize{}44.7} & {\footnotesize{}72.2} & {\footnotesize{}57.8} & {\footnotesize{}63.0} & {\footnotesize{}56.4} & {\footnotesize{}55.3} & {\footnotesize{}50.7} & {\footnotesize{}50.1} & {\footnotesize{}50.0} & {\footnotesize{}24.3} & {\footnotesize{}59.7} & {\footnotesize{}59.3} & {\footnotesize{}28.2} & {\footnotesize{}44.5} & {\footnotesize{}47.6}\tabularnewline
{\footnotesize{}CRN+MAC} & {\footnotesize{}72.8} & {\footnotesize{}56.7} & \textbf{\footnotesize{}84.5} & \textbf{\footnotesize{}71.7} & \textbf{\footnotesize{}75.9} & {\footnotesize{}70.5} & {\footnotesize{}76.2} & {\footnotesize{}90.7} & {\footnotesize{}75.9} & {\footnotesize{}57.2} & {\footnotesize{}76.1} & {\footnotesize{}92.8} & {\footnotesize{}91.0} & \textbf{\footnotesize{}87.4} & {\footnotesize{}85.4} & {\footnotesize{}75.8}\tabularnewline
\hline 
\textbf{\footnotesize{}OCRL+MA} & {\footnotesize{}77.4} & {\footnotesize{}56.7} & {\footnotesize{}81.2} & {\footnotesize{}64.6} & {\footnotesize{}65.0} & {\footnotesize{}90.0} & {\footnotesize{}93.4} & {\footnotesize{}90.1} & {\footnotesize{}77.0} & {\footnotesize{}93.5} & \textbf{\footnotesize{}77.8} & \textbf{\footnotesize{}92.9} & \textbf{\footnotesize{}91.3} & {\footnotesize{}82.5} & \textbf{\footnotesize{}89.5} & {\footnotesize{}77.8}\tabularnewline
\textbf{\footnotesize{}OCRL+LO} & \textbf{\footnotesize{}81.7} & \textbf{\footnotesize{}61.5} & {\footnotesize{}83.2} & {\footnotesize{}64.9} & {\footnotesize{}71.4} & \textbf{\footnotesize{}92.7} & \textbf{\footnotesize{}97.2} & \textbf{\footnotesize{}94.6} & \textbf{\footnotesize{}88.8} & \textbf{\footnotesize{}95.7} & {\footnotesize{}75.1} & {\footnotesize{}90.9} & {\footnotesize{}90.3} & {\footnotesize{}82.6} & {\footnotesize{}86.8} & \textbf{\footnotesize{}79.5}\tabularnewline
\hline 
\end{tabular}
\end{table*}

\subsection{Object-Centric vs. Grid Representation}

\begin{figure}
\begin{centering}
\includegraphics[width=1\columnwidth]{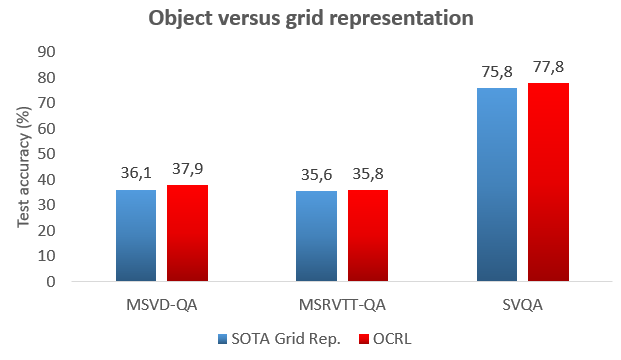}
\par\end{centering}
\caption{Object versus grid representation on MSVD-QA, MSRVTT-QA, and SVQA
datasets. Results for the grid representation is from \cite{le2020hierarchical,le2020neural-reason}.
For fair comparisons, MACNet is used as the reasoning module for all
approaches. \label{tab:Object-versus-patch-SVQA}}
\end{figure}

In order to understand the effects of the object-centric representation
learning (OCRL) in comparison with the popular grid representation
used by \cite{le2020neural-reason,le2020hierarchical}, we use a MACNet
\cite{hudson2018compositional} as the relational reasoning engine
for fair comparison. Fig.~\ref{tab:Object-versus-patch-SVQA} presents
comparison results on MSVD-QA, MSRVTT-QA, and SVQA datasets between
the \emph{OCRL} and non-object feature baselines in \cite{le2020neural-reason,le2020hierarchical}.
For the first two real datasets, the results confirm that the object-oriented
representation makes it a lot easier to arrive at correct answers
than the grid representation counterpart. In particular, the object-centric
representation outperforms other SOTA non-object methods on both MSVD-QA
and MSRVTT-QA (1.8 and 0.2 points improvement, respectively). For
the SVQA, our proposed OCRL model surpasses the grid representation
by roughly 2 points. Empirical results on the SVQA specifically reveal
the strong desire for proper object-centric representation towards
solving multi-step reasoning.

\subsection{Ablation Study\label{subsec:Ablation-study}}

We conduct ablation studies on the MSVD-QA dataset to justify the
contributions of different components in our model. Empirically, we
have noticed that the contribution of each component is more noticeable
when decreasing the number of reasoning steps of the relational module.
In particular, we set the number of reasoning steps of LOGNet as 2
in all ablated experiments. Results in Table~\ref{tab:Ablation-results}
reveal that each design component ameliorates the overall performance
of the model. The effects are detailed as follows:

\textbf{Position-specific mechanism:} We study the effect of the gating
mechanism in the Eq.~(\ref{eq:pos-app-feat}). In particular, we
replace this module with a simple concatenation. As can be seen, adding
gating mechanism leads to an improvement of 1.1 absolute points. The
results convincingly demonstrate the effects of our novel position-gated
mechanism comparing to the common approach of combining the appearance
and spatial feature \cite{le2020dynamic,wang2019neighbourhood}.

\textbf{Language-conditioned representation of temporal parts}: We
study the effect of the language on representing temporal parts in
Sec.~\ref{subsec:Language-conditioned-representat}. In particular,
we replace the temporal attention in Eq.~(\ref{eq:temporal_attention})
with a simple mean-pooling operation. We experience a significant
drop in performance from 37.6\% to 35.3\% when disregarding the language-condition
representation. We conjecture that the \emph{question-specific representations}
of object lives make it more ready for the later relational reasoning
to arrive at correct answers.

\textbf{Query-conditioned object graph:} We conduct a series of experiments
going from shallow GCNs to very deep GCNs to study the effects of
the temporal parts representation refinement based on the information
carried out by their surrounding objects. Empirical results suggest
that it gradually improves the performance when increasing the depth
of the GCN, and 6 layers are sufficient on the MSVD-QA dataset. 

\textbf{Object graphs summary:} We verify the effect of the sequential
modeling of temporal parts as described in Sec.~\ref{subsec:Dynamic-object-graphs}.
The simplest way of computing the summary of object graphs is via
an average-pooling operator over all elements in the sequence which
is to totally ignore the temporal dependencies between the elements.
Results in Table \ref{tab:Ablation-results} show that connecting
a chain of object temporal parts by a sequence model like BiLSTM is
beneficial. Comparing to the clip-based relation network \cite{le2020neural-reason},
a BiLSTM as in our model is less computationally expensive and easy
to implement.

\textbf{Contextual object: }Finally, we investigate the effect of
contextual feature in Eq.~(\ref{eq:GCN}). The performance of our
model notably decreases from 37.6 to 35.8 when removing this special
object. This result once again confirms the importance of the contextual
object in compensating for the loss of information caused by the imperfect
object detection.

\begin{table}
\caption{Ablation studies on the MSVD-QA dataset. ({*}) 6 GCN layers.\label{tab:Ablation-results}}

\centering{}%
\begin{tabular}{l|c}
\hline 
\multirow{1}{*}{Model} & \multicolumn{1}{c}{Validation Acc.}\tabularnewline
\hline 
\quad{}Default config. ({*}) & 37.6\tabularnewline
\hline 
\multicolumn{2}{l}{\textbf{Position-specific mechanism}}\tabularnewline
\hline 
\quad{}w/o Gating mechanism & 36.5\tabularnewline
\hline 
\multicolumn{2}{l}{\textbf{Language-conditioned rep. of temporal parts}}\tabularnewline
\hline 
\quad{}w/o Temporal attention & 35.3\tabularnewline
\hline 
\multicolumn{2}{l}{\textbf{Interaction graph of temporal parts}}\tabularnewline
\hline 
\quad{}w/ 1 GCN layer & 36.9\tabularnewline
\quad{}w/ 4 GCN layers & 37.3\tabularnewline
\quad{}w/ 8 GCN layers & 37.1\tabularnewline
\hline 
\multicolumn{2}{l}{\textbf{Object graphs summary}}\tabularnewline
\hline 
\quad{}w/o BiLSTM & 36.6\tabularnewline
\hline 
\textbf{Contextual object} & \tabularnewline
\hline 
\quad{}w/o Contextual Obj. & 35.8\tabularnewline
\hline 
\end{tabular}
\end{table}

\subsection*{}

\section{Discussion \label{sec:Discussion}}

We have introduced a novel neural architecture for object-centric
representation learning in video question answering. Object-centric
representation adds modularity and compositionality to neural networks,
offering structural alternatives to the default vectorial representation.
This is especially important in complex scenes in video with multi-object
dynamic interactions. More specifically, our representation framework
abstracts a video as an evolving relational graph of objects, whose
nodes and edges are conditionally inferred. We also introduce the
concept of r\a'esum\a'e that summarizes the live of an object over the entire
video. This allows seamless plug-and-play with existing reasoning
engines that operate on a set of items in response to natural questions.
The whole object-centric system is supported by a new dynamic graph
neural network, which learns to refine object representation given
the query and the context defined by other objects and the global
scene. Our architecture establishes new state-of-the-arts on MSVD-QA,
MSRVTT-QA, and SVQA \textendash{} the three well-known video datasets
designed for complex compositional questions, relational, spatial
and temporal reasoning.

{\small{}\bibliographystyle{plain}
\bibliography{egbib}
}{\small\par}
\end{document}